\documentclass{article}

\usepackage{arxiv}

\usepackage[utf8]{inputenc} % allow utf-8 input
\usepackage[T1]{fontenc}    % use 8-bit T1 fonts
\usepackage{hyperref}       % hyperlinks
\usepackage{url}            % simple URL typesetting
\usepackage{booktabs}       % professional-quality tables
\usepackage{amsfonts}       % blackboard math symbols
\usepackage{nicefrac}       % compact symbols for 1/2, etc.
\usepackage{microtype}      % microtypography
\usepackage{lipsum}		% Can be removed after putting your text content
\usepackage{graphicx}
\usepackage{natbib}
\usepackage{doi}

\title{Building Pre-train LLM Dataset for the Indic Languages: A Case
Study on Hindi}

\author{ Shantipriya Parida  \\
Silo AI, Helsinki, Finland \\
\texttt{shantipriya.parida@silo.ai} \\
	\And
	Shakshi Panwar \\
	Odia Generative AI, Banasthali Vidyapith, India\\
	\texttt{shakshipanwar942001@gmail.com} \\
 \And
Kusum Lata \\
Sharda University, India     \\
\texttt{kusumlata@nith.ac.in} \\
\And
Sanskruti Mishra \\
Vellore Institute of Technology, India\\
\texttt{sansi.1372@gmail.com} \\
\AND
Sambit Sekhar \\
Odia Generative AI\\
\texttt{sambitsekharai@gmail.com} \\
}

\renewcommand{\shorttitle}{\textit{Building Pre-train LLM Dataset for the Indic Languages: A Case
Study on Hindi} }

\hypersetup{
pdftitle={A template for the arxiv style},
pdfsubject={q-bio.NC, q-bio.QM},
pdfauthor={David S.~Hippocampus, Elias D.~Striatum},
pdfkeywords={First keyword, Second keyword, More},
}

\begin{document}
\maketitle

\begin{abstract}
	Large language models (LLMs) demonstrated transformative capabilities in many
applications that require automatically generating responses based on human instruction. However, the major challenge for building LLMs, particularly in Indic
languages, is the availability of high-quality data for building foundation LLMs.
In this paper, we are proposing a large pre-train dataset in Hindi useful for the Indic
language Hindi. We have collected the data span across several domains including
major dialects in Hindi. The dataset contains 1.28 billion Hindi tokens. We have
explained our pipeline including data collection, pre-processing, and availability
for LLM pre-training. The proposed approach can be easily extended to other Indic
and low-resource languages and will be available freely for LLM pre-training and
LLM research purposes.
\end{abstract}

% keywords can be removed

\section{Introduction}
A certain kind of language model known as pre-trained LLM has been trained on a sizable corpus of text data. Due to their capacity to allow machines to comprehend and produce human language, Language Models have grown in importance in the field of natural language processing (NLP) \cite{zhao2023survey}. Speech recognition, sentiment analysis, machine translation, audio-to-text conversion, and other tasks are all handled by NLP-based systems using language models. LLMs go through extensive training and fine-tuning, which can lead to increased dependability and performance. It is simpler to integrate them into current systems and applications because they frequently come with well-documented APIs and integration options\cite{naveed2023comprehensive}.

\begin{figure}[!htb]
    \begin{center}
    \includegraphics[scale=0.20]{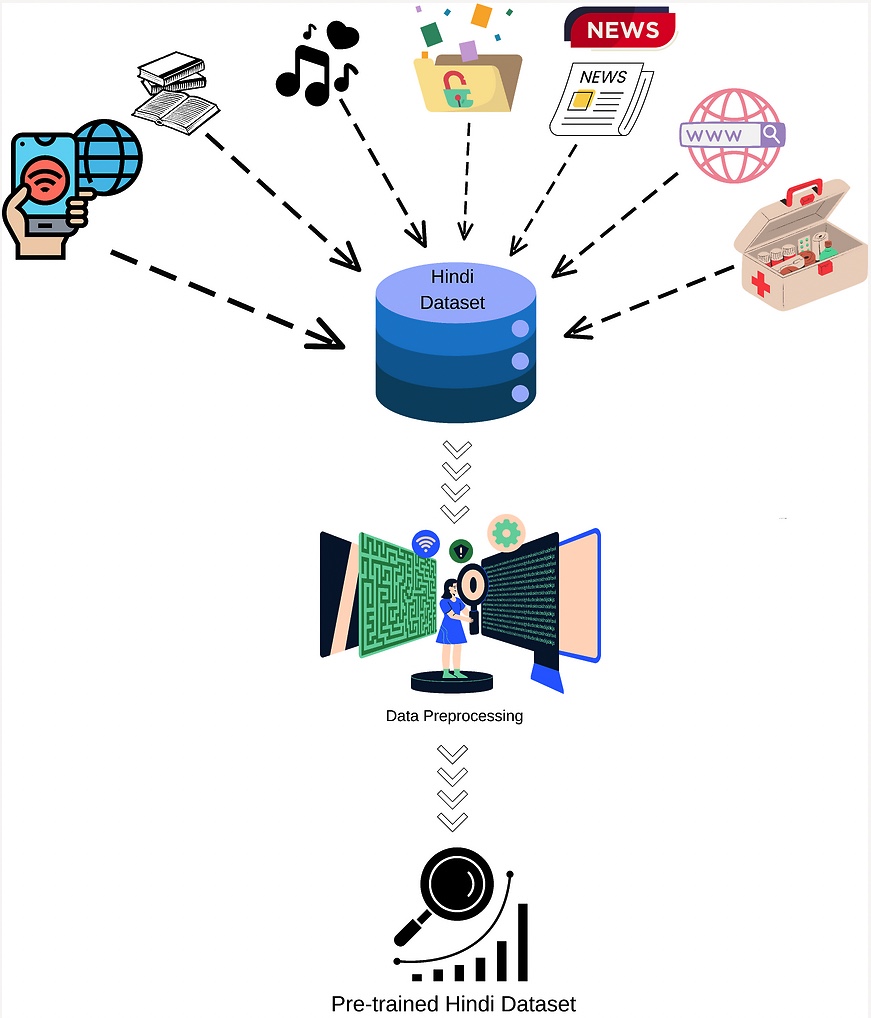} 
    \caption{The process of gathering information from several sources, such as news articles and supplementary texts, etc., and processing it to produce the pre-trained Hindi dataset is depicted in the figure.}
    \label{fig_overview}
    \end{center}
\end{figure}

Despite the fact that there are over 7,000 languages that are actively spoken worldwide, yet most NLP systems only employ a small number of languages, including English, Chinese, Urdu, Farsi, Arabic, French, and Spanish. Since many cultures are represented and inherited through different languages, there is an increasing interest in creating NLP for languages other than English. Text creation, summarization, and language understanding tasks have been impeded by the lack of strong pre-trained language models for non-English languages like Hindi, which has impeded the advancement of the development of NLP applications in these languages. Therefore, to support a variety of NLP applications and further the development of multilingual NLP, it is necessary to create more reliable and effective pre-trained language models for non-English languages. For this purpose, a large dataset is required for training models. 

With over 600 million speakers, Hindi is one of the most commonly spoken languages in the world. This poses both special potential and challenges for NLP research and applications. In contrast to languages such as English, there are still few large-scale, high-quality datasets and pre-trained language models available for Hindi. The lack of resources makes it difficult to create reliable NLP systems that are suited to the requirements of Hindi speakers. To tackle this difficulty, it is imperative to create big language models that have already been trained particularly for Hindi. These models can be used as the basis for many other NLP tasks, such as sentiment analysis, machine translation, text categorization, named entity identification, and more. Pre-trained language models can drastically cut down on the amount of data and processing power needed to create efficient NLP solutions in Hindi by utilizing the capabilities of deep learning and transfer learning techniques.

We aim to bridge this gap in this study by describing the procedure of creating big, pre-trained language models dataset for Hindi. We'll look at a variety of model architecture, training approaches, data-gathering methods, preprocessing methods, and assessment strategies that are specific to the Hindi language. Our objective is to facilitate breakthroughs in academic research as well as industrial applications by offering scholars and practitioners a thorough roadmap to creating cutting-edge NLP systems for Hindi.

\section{ Literature Survey}
%\label{sec:headings}
\label{gen_inst}
Our research article on the Hindi LLM pre-trained model dataset comprises an associated section where we examine specific investigations that make major contributions to the field of large language models and pre-trained model adaption to new languages. First off, \cite{zhao2023survey} offers a thorough summary of numerous big language models, covering their designs, approaches for training, and uses in a variety of languages. This study provides insightful information about the state of language modeling, which is essential background knowledge for comprehending the growth and developments in the area.
The important job of adapting pre-trained language models to new languages is addressed by \cite{csaki2023efficiently}; this difficulty is especially pertinent to our research on the Hindi LLM model. The study investigates effective methods for adjusting current models to manage the linguistic subtleties and traits of various languages.

The author \cite{joshi2022l3cube} introduces L3Cube-HindBERT and DevBERT pre-trained models, which symbolize particular advances in the context of Devanagari-based languages, such as Marathi and Hindi. 
This work is with our goal of creating a pre-trained model dataset specifically for the Hindi language since it emphasizes the significance of language-specific adaptations in meeting the needs of languages with distinctive scripts and linguistic traits. Altogether, these studies set the foundation and offer insightful information about large language models and how they are tailored to different linguistic circumstances, which helps us with our research on the Hindi LLM pre-trained dataset.
The MuRIL: Multilingual Representations for Indian Languages developed by \cite{khanuja2021muril} and discusses the drawbacks of the multilingual language models that are currently in use for Indian languages and presents MuRIL as an option for improvement. It points up problems, including subpar performance and handling transliteration and code-mixing complications, and emphasizes the necessity for models customized to India's linguistic variety. Although the work offers insightful information, it could be improved by future research directions and a deeper examination of the effects of code-mixing and transliteration.
Within the field of Hindi-specific pre-trained language models (PLMs), new developments have prompted the creation of HinPLMs \cite{huang2021hinplms}, which are intended to improve Hindi language natural language processing tasks. This paper captures the complex language patterns and semantics present in Hindi literature by utilizing large-scale corpora and advanced systems. Promising outcomes have also been observed in attempts to modify and enhance these models for particular tasks and domains, indicating their ability to tackle practical challenges in Hindi linguistic processing. 
The publication \cite{gala2024airavata} highlights a significant research vacuum, citing a widespread absence of support for Indian languages in contemporary language models, which are mostly focused on English. It highlights the crucial significance that different instruction-tuning datasets play in maximizing the efficacy of LLMs and calls attention to their inadequate availability for Hindi. The study emphasizes the value of creating an open ecosystem for additional research and innovation, and it shows how filling these gaps advances the discipline, especially in Indian languages.

Developing LLMs for high-resource languages like English has been the main focus of recent advances in NLP. Still, there is a long way to go in creating equivalent models for low- and mid-resource languages. In response to this problem, \cite{khan2024indicllmsuite} unveiled the INDICLLMSUITE, an extensive collection of tools created especially for the creation of Indic LLMs in 22 languages. Pre-training and fine-tuning datasets totaling 251 billion tokens and 74.8 million prompt-response pairs are included in the package. \cite{khan2024indicllmsuite} state that the resources have been carefully selected from a variety of sources, including websites, PDFs, videos, and translations. There are still several issues and difficulties even if the INDICLLMSUITE is a great effort in terms of offering resources for Indic language models. Assuring the correctness and dependability of carefully selected datasets requires quality assurance procedures, and increasing coverage to encompass other low-resource languages will improve inclusiveness.

\section{ Focused Language}
As a member of the Indo-Aryan branch of the Indo-European language family, Hindi is a highly organized language \cite{patil2008focus}. It is among India's official languages. The subject usually appears at the front of the sentence, followed by the object and the verb. This subject-object-verb (SOV) word order is what distinguishes the Hindi language structure. However, due to its flexible nature, Hindi allows for variations in word order to emphasize certain elements or express different complexities. Hindi has a large lexicon that incorporates words from many different languages, including English, Persian, Sanskrit, and Arabic. It uses the Devanagari script, a syllabic alphabet with unique letters for vowels and consonants.

In terms of grammar, Hindi is an extremely inflected language. Nouns are classified by gender (masculine or feminine), number (singular or plural), and case (nominative, accusative, dative, genitive, locative, etc.). Adjectives and pronouns share the same gender, number, and case as the nouns they modify. Hindi has a complex system of verb conjugation, with verbs inflected to indicate tense, aspect, mood, person, and number. Like Odia, Hindi makes use of postpositions rather than prepositions, which means that these markers are placed after the noun they govern. These postpositions play a crucial role in expressing various relationships and indicate concepts such as location, direction, possession, and time.

The process of building the pre-designed LLM dataset involved finding and compiling a variety of textual data from different domains, genres, and dialect variations in the Hindi language landscape. This includes literary works, historical archives, contemporary discourse, digital content, and other linguistic works that contribute to the richness and diversity of the Indian language corpus.

With a special focus on Hindi as the primary language of interest, the dataset provides a targeted approach to capture unique linguistic features, semantic subtleties, and contextual nuances prevalent in Hindi texts. This targeted research enables researchers and practitioners to develop language models and NLP algorithms that match the specific linguistic features and requirements of Hindi, thereby increasing their accuracy, flexibility, and applicability in real-world scenarios.

\section{Dataset Preparation}

 The initial
step involves obtaining various texts from different domains, which
include news articles, literary works, online content, and social media,
among others. It is essential to create a rich and representative dataset
that encapsulates the breadth of linguistic nuance and contextual
variation present in Hindi.

% \ref{sec:headings}.

\subsection{ Data Collection}
It is necessary to compile and select linguistic resources appropriate for Hindi language creation and understanding tasks to create datasets for the Hindi language that are specifically designed for pre-trained LLMs. %Building an optimal dataset for training a large language model (LLM) specifically for Hindi requires complex treatment. 
It includes a selection of datasets to improve effective pre-training on different content and linguistic nuances and different domains.

%\begin{figure}[!htb]
%    \begin{center}
%    \includegraphics[scale=0.15]{fig/fig_data_raw_data.jpeg} 
%    \caption{An Overview of Data Collection collected from a plethora of sources namely Wikipedia, Oscar, ai4bharat, song lyrics, etc.}
%    \label{fig_data_collection}
%    \end{center}
%\end{figure}

The first dataset, licensed by cc-by-sa-3.0 from \textit{Wikipedia }\cite{jones2017multilingual}, forms the basis of our corpus. With 43,670,526 characters spread across 1,850,408 sentences, this dataset provides a comprehensive summary of general knowledge and language use. The variety of topics and comprehensive coverage make it an invaluable resource for teaching language models for understanding and creating coherent texts in Hindi.

Complementing the \textit{Dialect Hindi Dataset} \cite{bafna2022empirical}, a repository that focuses on language change. This dataset adds 459,384 characters spread across 63,091 sentences, capturing the complexity of regional language nuances necessary for a robust language model. Covering the vernacular of 26 languages and dialects related to Hindi, the dataset provides valuable linguistic insight into regional variation, dialectal features, and cultural nuances.

Additionally, the \textit{AI4Bharat/IndicParaphrase dataset} \cite{kumar2022indicnlg}, licensed under cc-by-nc-4.0, adds another layer of linguistic diversity to our corpus. With 55,670,651 astounding numbers distributed across 5,864,552 sentences, this dataset provides a large set of sentences, expanding the dataset to include semantic variations and syntactic structures in Hindi. Completing translation expressions facilitates a deeper study of linguistic semantics and encourages the development of language models that can produce more diverse and contextually relevant results.

The \textit{Oscar dataset} \cite{suarez2019asynchronous,suarez2020monolingual}, available under the CC0-1.0 license, represents a major asset in developing a robust pre-built dataset that is transparently designed for LLM in Hindi. Boasting a whopping 745,990,971 words and containing 27,117,459 sentences, this dataset provides unparalleled linguistic data for comprehensive language model training.

The richness of \textit{Oscar dataset} extends beyond its size; and covers a variety of linguistic sources, including linguistic nuances, dialectical changes, and general thematic contexts in the use of Indian languages. This diversity allows for a deeper exploration and understanding of the complexities inherent in the Hindi language, improving the model's ability to understand and produce coherent texts across different linguistic domains and styles.

\begin{table}[!htb]
    \centering
    \small
    \begin{tabular}{c p{4cm} c c c}
        \toprule
        \textbf{S.no} & \textbf{Dataset} & \textbf{License} & \textbf{Total Token Count} & \textbf{Total Sentence Count} \\
        \midrule
        1 & Wikipedia Dataset & cc-by-sa-3.0 & 43.67 million & 1.85 million \\
        2 & Dialect Hindi Dataset & N/A & 0.46 million & 0.06 million \\
        3 & AI4Bharat IndicParaphrase & cc-by-nc-4.0 & 55.67 million & 5.86 million \\
        4 & Miracl Corpus & apache-2.0 & 33.66 million & 2.04 million \\
        5 & Oscar & cc0-1.0 & 745.99 million & 27.12 million \\
        6 & bigscience xP3all & apache-2.0 & 395.32 million & 21.86 million \\
        \midrule
        & \textbf{Total} & & \textbf{1.27 billion} & \textbf{58.79 million} \\
        \bottomrule
    \end{tabular}
    \caption{This table presents a comprehensive overview of pre-trained datasets, detailing key metrics such as token count, sentence count, and licensing information for datasets sourced from various sources including Wikipedia, ai4bharat, Miracl Corpus, Oscar, and others.}
    \label{tab:dataset_summary}
\end{table}

\textit{Miracl-Corpus} \cite{zhang2022making}, licensed by Apache-2.0, contributed significantly to our dataset by collecting 33,662,634 items consisting of 2,040,026 sentences. This corpus, licensed under the Apache License, provides a wealth of linguistic concepts and domain-specific content, adding valuable diversity to our dataset. Miracl-Corpus increases the breadth of topics and content represented in a pre-trained dataset with a large number and size of vocabulary, improving the model's understanding of the nuances of Indian languages in various domains and linguistic styles.

In addition, the addition of the \textit{big science/xP3all} \cite{muennighoff2022crosslingual} dataset further increases the linguistic diversity and domain coverage of our pre-trained dataset. Licensed under the Apache-2.0 license, this dataset contains 395,323,154 effective numbers distributed over 21,856,987 sentences. A number of linguistic variations, domain-specific terminology, and a variety of thematic content increase the flexibility of our dataset. By integrating Bigscience/xP3all, we equip our pre-trained models with language skills necessary for accuracy and adaptability to real-world problems, thereby improving their ability to understand and produce coherent texts in Hindi across multiple contexts and domains.

\begin{figure}[!htb]
    \begin{center}
    \includegraphics[scale=0.35]{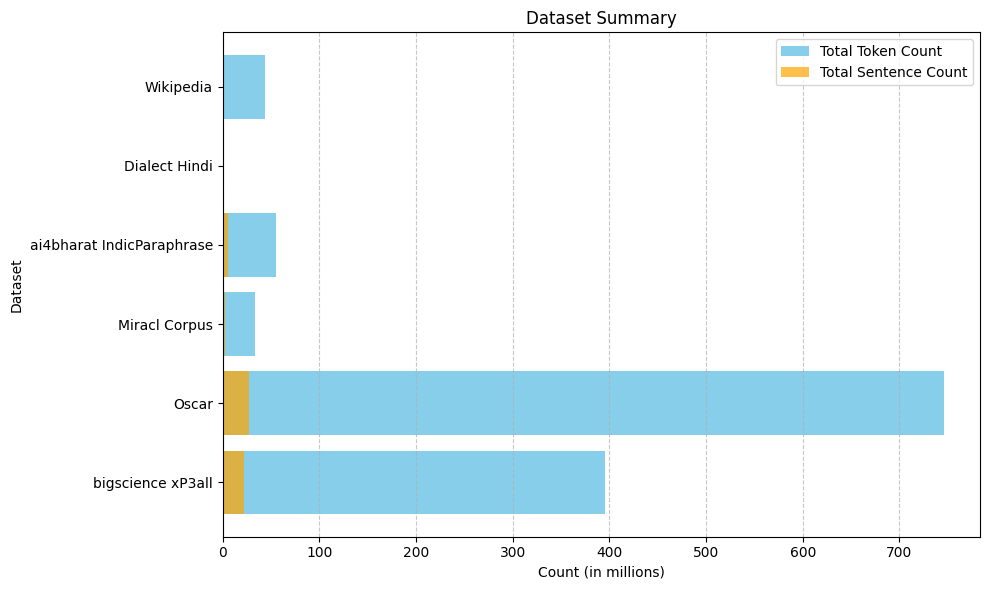} 
    \caption{This offers a thorough breakdown of the dataset statistics, covering measures like sentence and token counts that provide information regarding the scale and variety of the assembled datasets.}
    \label{fig_dataset_stat}
    \end{center}
\end{figure}

The statistics of the dataset are shown in Fig. \ref{fig_dataset_stat}.

%\begin{figure}[!htb]
%    \begin{center}
%    \includegraphics[scale=0.50]{fig/fig_bar_chart.png} 
%    \caption{Dataset Statistics}
%    \label{fig_dataset_stat}
%    \end{center}
%\end{figure}

This strategic integration allows our pre-engineered models to better understand the complexities of Hindi language usage, increasing its effectiveness in addressing diverse language and domain problems with accuracy. As a result, our pre-engineered dataset is designed as a reliable resource, fueling progress in natural language understanding and generation specifically for Indian languages.

By curating a dataset characterized by key figures and diverse content, our Hindi language corpus emerges as a powerful resource designed to facilitate LLM preparation specifically for Hindi.

%\begin{equation}
%	\xi _{ij}(t)=P(x_{t}=i,x_{t+1}=j|y,v,w;\theta)= {\frac {\alpha _{i}(t)a^{w_t}_{ij}\beta _{j}(t+1)b^{v_{t+1}}_{j}(y_{t+1})}{\sum _{i=1}^{N} \sum _{j=1}^{N} \alpha _{i}(t)a^{w_t}_{ij}\beta _{j}(t+1)b^{v_{t+1}}_{j}(y_{t+1})}}
%\end{equation}

%\subsubsection{Headings: third level}

%\paragraph{Paragraph}

\section{  Data Processing}

\label{sect_data_processing}

The initial stage of data processing begins with the exploration of different datasets, each presenting a unique set of challenges and opportunities. From Wikipedia's\cite{jones2017multilingual} knowledge repository to specialized corpora such as Hindi Dialect \cite{bafna2022empirical} and the ai4bharat IndicParaphrase dataset \cite{kumar2022indicnlg}, each source has added a unique flavor to the entire tapestry of linguistic information. In addition, the addition of datasets such as Miracl Corpus\cite{zhang2022making}, Oscar  \cite{suarez2019asynchronous},\cite{suarez2020monolingual}, and bigscience/xP3all \cite{muennighoff2022crosslingual} have added depth and breadth to the training corpus, enriching it with a variety of text genres, styles, and registers.

%\begin{figure}[!htb]
%    \begin{center}
%    \includegraphics[scale=0.30]{fig/fig_raw_dataset.png} 
%    \caption{Raw Dataset}
%    \label{fig_raw_dataset}
%    \end{center}
%\end{figure}

%\begin{figure}[!htb]
%    \begin{center}
%    \includegraphics[scale=0.30]{fig/fig_preprocessing.png} 
%    \caption{Processed Dataset}
%    \label{fig_preprocessing}
%    \end{center}
%\end{figure}

After the acquisition, the dataset was intensively developed to standardize its format and improve its suitability for training large-scale language models. This pre-processing begins by filtering out external metadata such as ID, URL, and all kinds of audio and compiling data containing relevant textual content. Special attention is paid to the regularization of text elements, including the removal of special characters, punctuation, and number artifacts, thereby creating a consistent and consistent corpus for efficient model training.

The culmination of this initial work resulted in the transformation of a raw dataset characterized by two main columns, \textit{filename} and \textit{content}, into a coherent and organized form. The former serves as a beacon, clarifying the provenance of each document in the corpus, while the latter contains the cleaned textual data content and format. This uniformity not only facilitates seamless integration across different datasets but also provides the basis for the development of robust LM models designed to understand and generate natural language text with exceptional fidelity.

In addition, advanced techniques such as language processing and domain-specific filtering are used to improve the quality and relevance of the training data. This includes identifying and correcting language-related errors, resolving transliteration issues, and filtering out domain-specific arguments or technical terms that may introduce interference or bias into the dataset.

The culmination of this initial work resulted in a refined and streamlined corpus characterized by uniformity, consistency, and relevance. The revised dataset, embellished with detailed textual content, forms the foundation for building an advanced LM model for India. In addition, the pre-processing step not only lays the groundwork for effective model training, but also facilitates downstream NLP tasks such as text classification, sentiment analysis, and machine translation.

\section{  Analysis and Discussion}
\label{sect_analysis}
\begin{figure}[!htb]
    \begin{center}
    \includegraphics[scale=0.35]{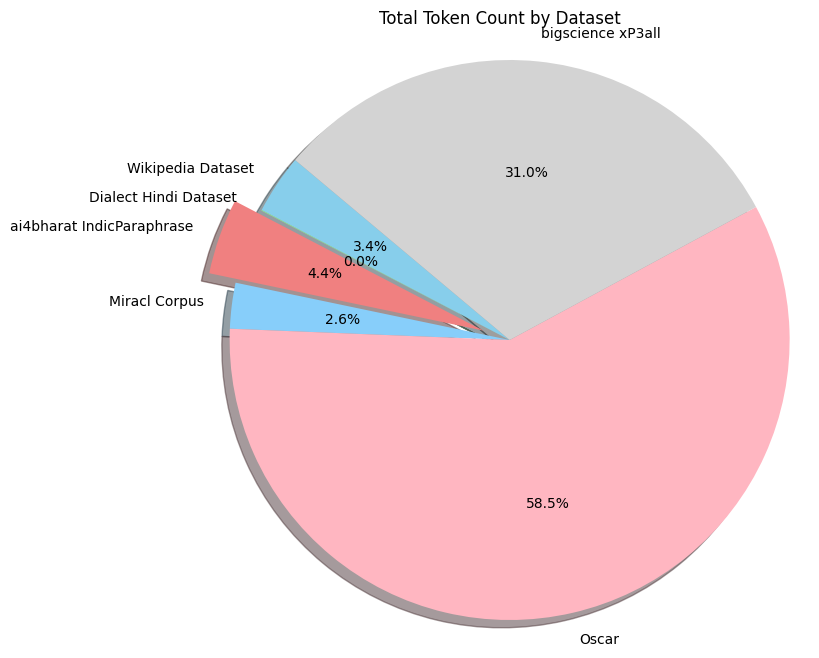} 
    \caption{This uses a pie chart to depict the distribution of distinct corpora collections, highlighting the proportionate representation of various datasets within the total corpus compilation.}
    \label{fig_data_collection}
    \end{center}
\end{figure}

%The distribution of the various collections of corpora is shown in the Fig. \ref{fig_data_collection}.
In the Analysis and Discussion section of this paper, we provide a comprehensive overview of various corpora collections used in the construction of pre-trained LM datasets. The distribution of these datasets is shown in Fig\ref{fig_data_collection}, which shows the number of tokens of language resources included in the unified dataset. Here, we explore findings, observations, and conclusions from working with this dataset that highlight challenges, opportunities, and contributions to the broader field of NLP for Indian languages, especially Indian languages.

One of the key observations is the wide range and variety of datasets collected to develop LLM for Indian languages. A total of 1.27 billion tokens collected in various datasets represent the wealth of linguistic resources available for developing complex language models that match the complexity of Hindi. This wealth of data provides an important opportunity to advance NLP research and practice in the Indian language landscape.

\subsection{ Domain Coverage}

\label{sect_domain}

%In exploring the domain environment of the dataset in question, it is important to note the unique data elements unique to each corpus. Each dataset contains unique linguistic content, contributing thematic richness and contextual significance to the wider context of language models and natural language processing.
As the collection of Hindi data spans multiple datasets covering various domains, in this section, we are highlighting the domains covered in our pre-trained Hindi LLM dataset. 

Starting with the \textit{Wikipedia dataset} \cite{jones2017multilingual}, uniqueness includes various general knowledge systems. It serves as a treasure trove of information on topics ranging from science and technology to history, culture, and more. The diversity of content in this dataset provides a broad coverage of the humanities, making it an invaluable resource for language comprehension problems.

%\begin{figure}[!htb]
%    \begin{center}
%    \includegraphics[scale=0.50]{fig/fig_bar_chart.png} 
%    \caption{Dataset Domain}
%    \label{fig_dataset_domain}
%    \end{center}
%\end{figure}

In contrast, the \textit{Hindi Dialect dataset (HinDialect)} \cite{bafna2022empirical} focuses on regional dialects and local language variations. We reveal the natural linguistic diversity in the dataset by examining the distribution of dialect features and linguistic landscape of Hindi-speaking regions and communities.

The HinDialect dataset, with its unique features and contribution to the broader goal of building a pre-built LLM dataset for Indian languages. This HinDialect dataset comprises folksongs from 26 Hindi-related languages and dialects forming a continuum in North India and nearby regions. These languages include \textit{Angika}, \textit{Awadhi}, \textit{Baiga}, \textit{Bengali}, \textit{Bhadrawahi}, \textit{Bhili}, \textit{Bhojpuri}, \textit{Braj}, \textit{Bundeli}, \textit{Chhattisgarhi}, \textit{Garhwali}, \textit{Gujarati}, \textit{Haryanvi}, \textit{Himachali}, \textit{Hindi}, \textit{Kanauji}, \textit{Khadi Boli}, \textit{Korku}, \textit{Kumaoni}, \textit{Magahi}, \textit{Malvi}, \textit{Marathi}, \textit{Nimadi}, \textit{Panjabi}, \textit{Rajasthani}, and \textit{Sanskrit}. The data, originally collected by the Kavita Kosh Project, features languages primarily spoken in North India, with Bengali also spoken in Bangladesh. Except for Korku, all languages are Indic, with most closely related to standard Hindi dialects genealogically.

The dataset exclusively uses the Devanagari script. For languages not typically written in Devanagari, such as Bengali and Gujarati, content has been transliterated by the Kavita Kosh Project. Each language's data is contained within a single text file, with folksongs separated by empty lines. The title of each folksong marks the beginning of a new piece, while line separation within folksongs is maintained.

By opening the \textit{AI4Bharat IndicParaphrase dataset} \cite{kumar2022indicnlg}, his unique contribution is based on paraphrase generation and linguistic equivalence. Designed to facilitate the identification of paraphrases, translations, and genealogical research, this dataset consists of pairs of sentences with the same meaning but different linguistic formulations. By including various paraphrastic transformations, it leads to advances in machine learning algorithms for natural language understanding and generation.

Likewise, the \textit{Miracle Corpus} \cite{zhang2022making} is distinguished by its particular focus on legal discourse and legislative terminology. Comprised of text from legal documents, court cases, and administrative records, it provides domain-specific resources for legal domain applications. The datasets unique content provides valuable insight into legal language usage that facilitates the development of domain-specific language models and text analysis tools.

In addition, the \textit{Oscar dataset} \cite{suarez2019asynchronous,suarez2020monolingual} is distinguished by its eclectic collection of texts from various media types and sources. It covers a wide range of textual content, from literary works and academic publications to news articles and social media posts. This diversity allows language models trained on the Oscar dataset to demonstrate robustness and adaptability across different communication domains and styles.

Finally, the \textit{Big Science/xP3all dataset} \cite{muennighoff2022crosslingual} focuses on scientific concepts, technical terminology, and research-related discourse. It serves as a comprehensive repository of scientific knowledge gathered from scientific literature, academic publications, and technical documents. This dataset enables language models to understand and generate text about complex scientific topics, improving the understanding and communication of science in natural language.

%\begin{figure}[!htb]
%   \begin{center}
%    \includegraphics[scale=0.50]{fig/fig_word_cloud.png} 
 %   \caption{Bringing the Content to Life: A Word Cloud Display of the Dataset Description}
  %  \label{fig_word_cloud}
   % \end{center}
%\end{figure}

Each data set contributes unique data elements that enrich the thematic scope and contextual depth of the trained language model. By highlighting the distinctive features of each dataset, we gain a deeper understanding of their contribution to the development of language models and natural language processing technology specific to Indian languages.

\section{  Use Cases}

\label{sect_usecases}

Integrating various datasets into a single corpus is a major task in the effort to build a pre-train LLM dataset for the Hindi language. This comprehensive dataset, carefully curated from various sources, contains a harmonious combination of linguistic diversity, cultural richness, and thematic depth, thus providing an invaluable resource for developing NLP capabilities in the Indian language landscape.

\paragraph{Pre-training:} Large datasets are the main source for training LLM for Indian languages, especially Hindi language. By exposing the model to a variety of linguistic data, the pre-training process allows the model to learn linguistic patterns, semantic relationships, and language-specific contextual nuances. These pre-trained models can be well adapted to downstream tasks such as sensory analysis, text generation, and machine translation \cite{mujadia2023assessing}, thus improving performance and adaptability.

\paragraph{Language Model:} The comprehensive dataset facilitates the development of robust language models tailored specifically for Hindi and other Indian languages. The dataset's extensive linguistic information may be used for language modeling tasks including text generation, sentence completion, and next-word prediction. By training language models on this corpus, researchers and developers can improve the model's understanding of Indian language structure, vocabulary, and semantics, making texts more accurate and contextually relevant.

\paragraph{Generating synthetic data:} The variety of domains, genres, and dialect variations represented in the dataset makes it an ideal source for generating synthetic data \cite{article}. Synthetic data generated from this corpus can be used to augment existing datasets, solve data scarcity problems, and increase the power of machine learning models. Applications of synthetic data generation include data augmentation to train classifiers, sentence structure generation for data diversity, and generation of artificial training examples for rare linguistic phenomena.
\paragraph{Domain-Specific Improvement:} The dataset covers many domains like literature, history, science, technology, and more. Researchers and practitioners can use this domain diversity to tailor pre-engineered language models for specific application domains. For example, in domain-specific tasks, language models can perform better when fed domain-specific data from a unified corpus, such as dataset fine-tuning, legal document analysis, medical text understanding, and financial statement summarization.
\paragraph{Multilingual NLP Research:} Apart from Hindi, the dataset can support multilingual NLP related to other Indian languages. By extending pre-training and fine-tuning methods to languages such as Bengali, Tamil, Telugu, and Gujarati, researchers can create language models and NLP programs that match the linguistic diversity of the Indian subcontinent. This promotes inclusiveness and accessibility in NLP research and application in various linguistic communities.

In total, \textit{1.28 billion} tokens serve as a foundation for research, innovation, and application in natural language processing for Indian languages, contributing to the development of language technologies that address the unique linguistic needs and challenges of the region.

In addition, the unified dataset serves as a convenient place to facilitate domain-specific customization of language models, so that researchers and practitioners use LLMs for specific applications and situations. By fine-tuning pre-trained models in domain-specific datasets derived from integrated corpora, stakeholders can improve the performance and effectiveness of language models in tasks such as sentiment analysis, document summarization, identity recognition, and intent classification, thereby improving their applicability in a real scenario. The aggregated data sets help to better model the language for downstream tasks such as sentiment analysis, machine translation, summarizing, and question answering. By fitting pre-trained models to domain-specific datasets derived from integrated corpora, researchers can tune language models to excel in specific applications, thereby improving performance and adaptability in real-world scenarios.

\section{  Availability}

The collected and processed data %(Figure \ref{fig_data_collection})  
are available through the Hugging Face:

\url{https://huggingface.co/datasets/Hindi-data-hub/odaigen_hindi_pre_trained_sp}

%repository \nolinkurl{Hindi-data-hub/odaigen_hindi_pre_trained_sp} of hindi pre-trained LLM dataset \nolinkurl{https://huggingface.co/datasets/Hindi-data-hub/odaigen_hindi_pre_trained_sp}. 

Researchers and practitioners can freely access the dataset for practice, model training, and further research.

\section{ Conclusion and Futurework}
\label{sect_conclusion}
In this research, we describe a complete strategy to addressing the issues of developing large pre-trained language models (LLMs) for Hindi, with an emphasis on data collecting, preprocessing, and availability. By collecting a broad and extensive dataset of 1.28 billion tokens from multiple sources, topics, and dialects, we have created the groundwork for furthering the field of natural language processing (NLP) studies as well as applications. the paper Building a Pre-Trained LLM Dataset for Hindi Language is an important step in the development of a Hindi-focused, large-scale, Hindi-specific pre-trained language model (LLM). Comprehensive integration of various datasets and intelligent preprocessing efforts lay the foundation for producing reliable and context-aware LLMs. Despite its contribution, the study recognizes some limitations and areas for future research.

The provided dataset provides a comprehensive library of linguistic assets that encapture Hindi's particular subtleties, cultural complexity, and thematic depth. We have turned unstructured textual information into a refined dataset sufficient for training strong and context-aware LLMs built specifically for Hindi. The dataset's domain coverage includes a wide range of themes, genres, and language variances namely different dialects and nuances of the vernacular language, allowing for the creation of models that are capable of understanding and producing coherent texts across a variety of fields and styles.

%\begin{figure}[!htb]
%    \begin{center}
%        \includegraphics[scale=0.30]{fig/fig_future_work.png} 
%        \caption{Defining Future Directions for Natural Language Processing Research: Improving Evaluation Tools, Encouraging Usability, and Growing Uses of Pre-Trained Language Models in Hindi and Other Languages.}
%        \label{fig_future_work}
%    \end{center}
%\end{figure}

In addition, future work can be explored to refine evaluation criteria and indicators to more comprehensively assess LLM performance. Standardized evaluation procedures and clear benchmarks are needed to effectively compare different models and approaches. By creating a reliable assessment framework, researchers can gain a deeper understanding of the strengths and weaknesses of LLM trainees, thereby leading to continuous improvement and innovation in the field.

Future research can explore ways to increase the availability and accessibility of resources for the wider research community. Facilitating easy access to datasets, source code, and pre-engineered models can promote collaboration, accelerate research, and democratize access to advanced language processing technologies. In addition, initiatives to support multilingualism and inclusiveness in the development of language models can contribute to digital inclusion and linguistic opportunities in diverse language communities.

\section{  Limitation}

\label{sect_limitation}

While the huge pre-trained LLM dataset described in this study provides substantial prospects for furthering the resources available for research opportunities, we need to acknowledge the limitations associated.
Despite attempts to pre-process data, there may still be flaws or inconsistencies that affect the efficacy of trained models. Furthermore, biases inherent in the source data, such as cultural or representational biases, may impact the action patterns of language models, resulting in unforeseen effects in downstream applications.
\begin{itemize}
\item
\textbf{Data Accuracy and Prejudice}
The paper recognizes the complexity of data bias and representation inherent in structured datasets. Despite careful efforts to collect diverse linguistic samples, bias can persist which can lead to misrepresentation or misrepresentation of certain linguistic and cultural aspects. These constraints may impede the generalization and implementation of trained LLMs, particularly in real-world contexts characterized by linguistic diversity, because structured datasets may provide insufficient coverage of linguistic change. 

\item 
\textbf{Addressing Linguistic Diversity and Evaluation Criteria}
Although this paper emphasizes the importance of linguistic diversity, it may not be sufficiently represented in some dialects, registers, or domains. As a result, LLMs trained in such information may struggle to accurately understand and produce texts across all the linguistic variations that exist in the Indian language landscape.
In addition, the paper can benefit from further clarification on the evaluation criteria and indicators used to evaluate LLM performance. A standardized evaluation procedure is needed to compare the effectiveness of different models and approaches in the field. Clear standards will allow you to understand the strengths and shortcomings of an LLM.

\item 
\textbf{Language Diversity and Domain Coverage}
While the dataset seeks to capture a wide range of linguistic content, it may not fully represent the breadth of language. Certain dialects, vernacular nuances, or specialized domains may be undervalued or absent from the dataset, restricting the applicability of trained models to a wide range of linguistic settings. Furthermore, the dataset's inclusion of specific areas or topics may be unequal, thereby biasing model performance.

\item 
\textbf{Resource Constraints in Developing}
The development and upkeep of a large pre-trained LLM dataset necessitate enormous resources for data gathering, preparation, storage, and computational infrastructure. While attempts have been undertaken to put together and consolidate the dataset, resource restrictions may limit its scalability and accessibility to researchers as well as practitioners, especially those working in resource-constrained situations.

\item 
\textbf{Ethical and Privacy Considerations}
The collection and use of huge amounts of textual data raises serious ethical and privacy concerns, specifically about user authorization, data anonymization, and the potential exploitation of sensitive information. While attempts have been made to follow ethical norms and data protection guidelines, extra caution and transparency in managing and securing the information is required to guarantee that privacy rights and ethical values are upheld.

\end{itemize}

\bibliographystyle{unsrtnat}
\bibliography{references}  %%% Uncomment this line and comment out the ``thebibliography'' section below to use the external .bib file (using bibtex) .





% \end{thebibliography}

\end{document}